# Equivalence of Two Expressions of Principal Line

Cheng-Yen Hsu, Hsin-Yi Chen and Jen-Hui Chuang

*Abstract*—Geometry-based camera calibration using principal line is more precise and robust than calibration using optimization approaches; therefore, several researches try to re-derive the principal line from different views of 2D projective geometry to increase alternatives of the calibration process. In this report, algebraical equivalence of two expressions of principal line, one derived w.r.t homography and the other using for two sets of orthogonal vanishing points, is proved. Moreover, the extension of the second expression to incorporate infinite vanishing point is carried out with simple mathematics.

## I. INTRODUCTION

Geometry-based camera calibration is more precise and robust than traditional calibration approaches based on algebraic optimization [1-3] due to the consideration of camera parameters individually. Chen [4] exploited geometric relationship between a vanishing line on image plane and a plane which contains optical center and principal point (PP), and is perpendicular to object plane, to sequentially derive three rotation angles, three translation distances and a focal length. However, the use of non-planar calibrated target, limit its usage. Chuang et al. [5-7] consider similar geometry but formally define principal line (PL) as the axis of symmetry of the 3D plane containing a checkerboard pattern (CB plane), i.e., a symmetric image pattern corresponds to a symmetric pattern on the CB plane.

Fig. 1 illustrates the geometry of a PL, wherein both *x*-axis and *X*-axis are parallel to the intersection of the two plane and *y*-axis correspond to the PL. As each PL thus defined for each camera-pattern pose will pass through the camera principal point (PP), the PP can then be estimated by finding the intersection of linearly independent PLs, possibly from a least square solution. Moreover, outliers in checkerboard patterns can also be identified easily in the early stage of calibration and removed from the subsequent estimation of other camera parameters.

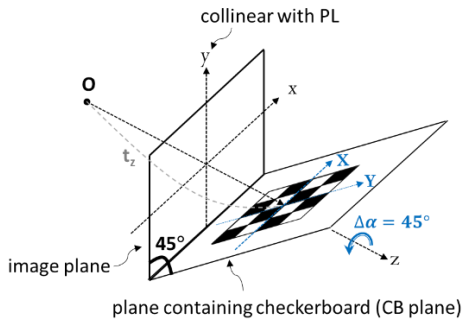

**Figure 1.** The geometry of a PL.

Wang et al. [8] rederive the same closed-from solution of PL for light field camera calibration. The same geometric relationship among image plane, a rectangle plane, and PL is considered with a different projection model, i.e., a new homography is derived according to projection model of the standard light field camera. Although the derivation of PL starts from line equations of four edges of a rectangular pattern, the final results are exactly the same as [7]. Higher accuracy is also demonstrated with the final results obtained from their PL-based calibration algorithm.

In [9], Yang and Zhao derive a different closed-from expression of PL for orthogonal vanishing points. Without knowing the homography between image plane and CB plane, as in [7], two sets of orthogonal vanishing points projected from the 3D space are used for PL derivation, with experimental results showing comparable accuracy in the calibration results. However, no proof of the equivalence of two closed-form expressions of PL are elaborated in [9]. Moreover, the expression in [9] is derived for *finite* orthogonal vanishing points (nonparallel image lines), while there is no such constraint in [7].

In this report, to address the above issues, the two PL expressions are first reviewed in Sec. II, followed by the proof of their equivalence in Sec. III. Then, simple extensions of the PL expression in [9] for *infinite* orthogonal vanishing points are provided in Sec. IV, before some concluding remarks are given in Sec. V.

## II. RELATED WORK

The principal line introduced in [7] plays an important role in the camera calibration, as a PL can be obtained for each checkerboard pattern and will pass through the principal point of the camera. Therefore, the PP can be determined as the intersection of a set of linearly independent PLs. Based on mathematical derivations provided in [7], if the homographic relationship between a CB plane and the image plane is represented by the homography matrix,

$$H = \begin{bmatrix} h_1 & h_2 & h_3 \\ h_4 & h_5 & h_6 \\ h_7 & h_8 & h_9 \end{bmatrix} \quad (1)$$

the principal line can be expressed as

$$(-h_1 h_8 + h_2 h_7)u + (-h_4 h_8 + h_5 h_7)v + c = 0, \quad (2)$$

where $c =$
$$-\frac{(h_2^2 + h_5^2 - h_1^2 - h_4^2)h_7 h_8 + (h_1 h_2 + h_4 h_5)(h_7^2 - h_8^2)}{h_7^2 + h_8^2}$$

On the other hand, a different expression of PL is also derived in [9] based on orthogonal vanishing points (OVPs) of a planar scene. In particular, given the first set of two OVPs ($P_{v1} = (m_1, n_1, 1)$ and $P_{v2} = (m_2, n_2, 1)$), and the second set ($P_{v3} = (m_3, n_3, 1)$, and $P_{v4} = (m_4, n_4, 1)$), the PL can be expressed as

$$(m_2 - m_1)u + (n_2 - n_1)v + c = 0, \quad (3)$$

where $c = -[(m_2 - m_1)\dfrac{m_1 m_2 - m_3 m_4}{m_1 + m_2 - m_3 - m_4} + (n_2 - n_1)\dfrac{n_1 n_2 - n_3 n_4}{n_1 + n_2 - n_3 - n_4}]$

In this report, the equivalence of the two expressions of PL, i.e., Equations (2) and (3), will first be derived in the next section, for two sets of (finite) orthogonal vanishing points of two squares on the CB plane, as shown in Fig. 2, following procedure outlined in [9] for a homographic transformation between image plane and CB plane. Similar derivations for more general situations will be considered in Sec. IV.

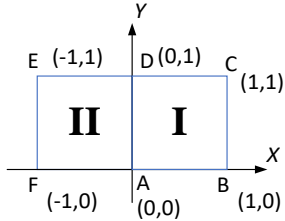

**Figure 2.** Two unit squares used in this report to prove the equivalence of (2) and (3).

### III. SHOWING THE EQUIVALENCE OF (2) AND (3)

*A. Obtaining $P_{v1}$ and $P_{v2}$*

Consider the homogeneous coordinates of the four vertices of Square I, i.e., $A = (0,0,1)$, $B = (1,0,1)$, $C = (1,1,1)$, and $D = (0,1,1)$. Based on the homography matrix given in (1), the homogeneous coordinates of images of these vertices can be expressed as

$$A' = \begin{bmatrix} \dfrac{h_3}{h_9} \\ \dfrac{h_6}{h_9} \\ 1 \end{bmatrix}, \quad B' = \begin{bmatrix} \dfrac{h_1 + h_3}{h_7 + h_9} \\ \dfrac{h_4 + h_6}{h_7 + h_9} \\ 1 \end{bmatrix},$$

$$C' = \begin{bmatrix} \dfrac{h_1 + h_2 + h_3}{h_7 + h_8 + h_9} \\ \dfrac{h_4 + h_5 + h_6}{h_7 + h_8 + h_9} \\ 1 \end{bmatrix}, \quad D' = \begin{bmatrix} \dfrac{h_2 + h_3}{h_8 + h_9} \\ \dfrac{h_5 + h_6}{h_8 + h_9} \\ 1 \end{bmatrix}. \quad (4)$$

In this subsection, it is assumed that $P_{v1}$ is the intersection of the $\overleftrightarrow{A'B'}$ and $\overleftrightarrow{C'D'}$ and $P_{v2}$ is the intersection of the $\overleftrightarrow{A'D'}$ and $\overleftrightarrow{B'C'}$. For $P_{v1}$, we first calculate the directional vector

$$\overrightarrow{A'B'} = \begin{bmatrix} \dfrac{h_1 h_9 - h_3 h_7}{h_9(h_7 + h_9)} \\ \dfrac{h_4 h_9 - h_6 h_7}{h_9(h_7 + h_9)} \end{bmatrix}. \quad (5)$$

Assume $\overrightarrow{A'B'}$ exists, both $h_9$ and $(h_7+h_9)$ will be nonzero, and we have

$$\overrightarrow{A'B'} = \begin{bmatrix} h_1 h_9 - h_3 h_7 \\ h_4 h_9 - h_6 h_7 \end{bmatrix}, \quad (6)$$

and the line equation of $\overleftrightarrow{A'B'}$ can be expressed as

$$(h_4 h_9 - h_6 h_7)u + (-h_1 h_9 + h_3 h_7)v + Constant1 = 0 \quad (7)$$

By substituting $A'$ in equation (4) into (7), we will have

$$Constant1 = -h_3 h_4 + h_1 h_6. \quad (8)$$

Similarly, the line equation of $\overleftrightarrow{C'D'}$ can be expressed as

$$(h_5 h_7 + h_6 h_7 - h_4 h_8 - h_4 h_9)u$$
$$+(-h_2 h_7 - h_3 h_7 + h_1 h_8 + h_1 h_9)v + Constant2 = 0. \quad (9)$$

By substituting $D'$ in equation (4) into (9), we will have

$$Constant2 = h_2 h_4 + h_3 h_4 - h_1 h_5 - h_1 h_6 \quad (10)$$

Therefore, $P_{v1}$ can be obtained as the intersection of the two lines represented by (7) and (9), e.g., by evaluating the outer product

$$P_{v1} = \overrightarrow{A'B'} \times \overrightarrow{C'D'} = \begin{vmatrix} \vec{i} & \vec{j} & \vec{k} \\ a_1 & b_1 & c_1 \\ a_2 & b_2 & c_2 \end{vmatrix}, \quad (11)$$

where $a_i$, $b_i$, and $c_i$ are the coefficients in (7) and (9). Then, the three components in (11) can be evaluated as

$$\begin{aligned} M_1 &= b_1 c_2 - b_2 c_1 \\ &= h_1(-h_2 h_4 h_9 + h_2 h_6 h_7 - h_1 h_6 h_8 \\ &\quad + h_1 h_5 h_9 - h_3 h_5 h_7 + h_3 h_4 h_8) \\ &= h_1 \cdot \det(H), \end{aligned} \quad (12)$$

$$\begin{aligned} N_1 &= c_1 a_2 - c_2 a_1 \\ &= h_4(-h_2 h_4 h_9 + h_2 h_6 h_7 - h_1 h_6 h_8 \\ &\quad + h_1 h_5 h_9 - h_3 h_5 h_7 + h_3 h_4 h_8) \\ &= h_4 \cdot \det(H), \end{aligned} \quad (13)$$

and

$$\begin{aligned} k_1 &= a_1 b_2 - a_2 b_1 \\ &= h_7(-h_2 h_4 h_9 + h_2 h_6 h_7 - h_1 h_6 h_8 \\ &\quad + h_1 h_5 h_9 - h_3 h_5 h_7 + h_3 h_4 h_8) \\ &= h_7 \cdot \det(H), \end{aligned} \quad (14)$$

or, if $k_1 \neq 0$,

$$P_{v1} = (m_1, n_1, 1) = \left(\frac{h_1}{h_7}, \frac{h_4}{h_7}, 1\right). \tag{15}$$

Note that if no three vertices of Square I are colinear, we will have $\det(H) \neq 0$ as the homographic transformation is bijection. Besides, if $P_{v1}$ does not locate at infinity we will have $h_7 \neq 0$.

As for obtaining the location of $P_{v2}$, the exact procedure used to derive (7) and (9) can be employed, and line equations of $\overleftrightarrow{A'D'}$ and $\overleftrightarrow{B'C'}$ can be found as

$$(h_4 h_9 - h_6 h_7)u + (-h_1 h_9 + h_3 h_7)v - h_3 h_5 + h_2 h_6 = 0 \tag{16}$$

and

$$\begin{aligned}(h_5 h_7 + h_5 h_9 - h_4 h_8 - h_6 h_8)u \\ +(-h_2 h_7 - h_2 h_9 + h_1 h_8 + h_3 h_8)v \\ -h_1 h_5 - h_3 h_5 + h_2 h_4 + h_2 h_6 = 0,\end{aligned} \tag{17}$$

respectively.

Thus, the three components of the intersection of $\overleftrightarrow{A'D'}$ and $\overleftrightarrow{B'C'}$ can be obtained, similar to that formulated in (11), as

$$\begin{aligned}M_2 &= h_2(h_1 h_5 h_9 - h_2 h_4 h_9 + h_3 h_4 h_8 \\ &\quad - h_3 h_5 h_7 + h_2 h_6 h_7 + h_1 h_6 h_8) \\ &= h_2 \cdot \det(H),\end{aligned} \tag{18}$$

$$\begin{aligned}N_2 &= h_5(-h_3 h_5 h_7 + h_3 h_4 h_8 + h_2 h_6 h_7 \\ &\quad + h_1 h_5 h_9 - h_2 h_4 h_9 - h_1 h_6 h_8) \\ &= h_5 \cdot \det(H),\end{aligned} \tag{19}$$

and

$$\begin{aligned}k_2 &= h_8(h_1 h_5 h_9 + h_2 h_6 h_7 - h_1 h_6 h_8 \\ &\quad - h_2 h_4 h_9 - h_3 h_5 h_7 + h_3 h_4 h_8) \\ &= h_8 \cdot \det(H),\end{aligned} \tag{20}$$

Or, if $k_2 \neq 0$,

$$P_{v2} = (m_2, n_2, 1) = \left(\frac{h_2}{h_8}, \frac{h_5}{h_8}, 1\right). \tag{21}$$

B. *Obtaining $P_{v3}$ and $P_{v4}$*

Consider Square II in Fig. 2, and homogeneous coordinates of its two vertices $E = (-1,1,1)$ and $F = (-1,0,1)$. Based on the homography matrix in (1), the homogeneous coordinates of images of these two vertices can be expressed as

$$E' = \begin{bmatrix} \frac{-h_1 + h_2 + h_3}{-h_7 + h_8 + h_9} \\ \frac{-h_4 + h_5 + h_6}{-h_7 + h_8 + h_9} \\ 1 \end{bmatrix}, \quad F' = \begin{bmatrix} \frac{-h_1 + h_3}{h_7 + h_9} \\ \frac{-h_4 + h_6}{h_7 + h_9} \\ 1 \end{bmatrix}. \tag{22}$$

Therefore, another pair of orthogonal vanishing points can be identified as the intersection of $\overleftrightarrow{A'C'}$ and $\overleftrightarrow{D'F'}$ ($P_{v3}$) and the intersection of $\overleftrightarrow{A'E'}$ and $\overleftrightarrow{B'D'}$ ($P_{v4}$).

By procedure adopted earlier to find other line equations, line equations of $\overleftrightarrow{D'F'}$ and $\overleftrightarrow{A'C'}$ can be found as

$$\begin{aligned}(-h_5 h_7 + h_5 h_9 - h_6 h_7 + h_4 h_8 + h_4 h_9 - h_6 h_8)u \\ +(-h_1 h_8 - h_1 h_9 + h_3 h_8 + h_2 h_7 - h_2 h_9 + h_3 h_7)v \\ +(-h_2 h_4 + h_2 h_6 - h_3 h_4 + h_1 h_5 + h_1 h_6 - h_3 h_5) = 0\end{aligned} \tag{23}$$

and

$$\begin{aligned}(h_6 h_7 + h_6 h_8 - h_4 h_9 - h_5 h_9)u \\ +(h_1 h_9 + h_2 h_9 - h_3 h_7 - h_3 h_8)v \\ +(h_3 h_4 + h_5 h_5 - h_1 h_6 - h_2 h_6) = 0,\end{aligned} \tag{24}$$

respectively, and the outer product in (11) can again be adopted to find the three components of $P_{v3}$ as

$$\begin{aligned}M_3 &= -(h_1 + h_2)(h_1 h_5 h_9 + h_2 h_6 h_7 - h_1 h_6 h_8 \\ &\quad + h_2 h_4 h_9 - h_3 h_5 h_7 + h_3 h_4 h_8) \\ &= -(h_1 + h_2) \cdot \det(H),\end{aligned} \tag{25}$$

$$\begin{aligned}N_3 &= -(h_4 + h_5)(h_1 h_5 h_9 + h_2 h_6 h_7 - h_1 h_6 h_8 \\ &\quad - h_2 h_4 h_9 - h_3 h_5 h_7 + h_3 h_4 h_8) \\ &= -(h_4 + h_5) \cdot \det(H),\end{aligned} \tag{26}$$

and

$$\begin{aligned}k_3 &= -(h_7 + h_8)(h_1 h_5 h_9 + h_2 h_6 h_7 - h_1 h_6 h_8 \\ &\quad - h_2 h_4 h_9 - h_3 h_5 h_7 + h_3 h_4 h_8) \\ &= -(h_7 + h_8) \cdot \det(H),\end{aligned} \tag{27}$$

or

$$P_{v3} = (m_3, n_3, 1) = \left(\frac{h_1 + h_2}{h_7 + h_8}, \frac{h_4 + h_5}{h_7 + h_8}, 1\right). \tag{28}$$

Similar to the derivation of (23) and (24), we can obtain line equations of $\overleftrightarrow{A'E'}$ and $\overleftrightarrow{B'D'}$ as

$$\begin{aligned}(h_4 h_9 + h_6 h_8 - h_6 h_7 - h_5 h_9)u \\ +(h_3 h_7 + h_2 h_9 - h_1 h_9 - h_3 h_8)v \\ +(h_1 h_6 + h_3 h_5 - h_3 h_4 - h_2 h_6) = 0\end{aligned} \tag{29}$$

and

$$\begin{aligned}(h_4 h_8 + h_4 h_9 + h_6 h_8 - h_5 h_7 - h_5 h_9 - h_6 h_7)u \\ +(h_2 h_7 + h_2 h_9 + h_3 h_7 - h_1 h_8 - h_1 h_9 - h_3 h_8)v \\ +(h_1 h_5 + h_1 h_6 + h_3 h_5 - h_2 h_4 - h_2 h_6 - h_3 h_4) = 0,\end{aligned} \tag{30}$$

respectively, and the three components of $P_{v4}$ as

$$\begin{aligned}M_4 &= (h_1 - h_2)(h_2 h_6 h_7 - h_1 h_6 h_8 - h_3 h_5 h_7 \\ &\quad + h_3 h_4 h_8 + h_1 h_5 h_9 + h_2 h_4 h_9) \\ &= (h_1 - h_2)(S) = (h_1 - h_2) \cdot \det(H),\end{aligned} \tag{31}$$

$$\begin{aligned}N_4 &= (h_4 - h_5)(h_1 h_5 h_9 - h_2 h_4 h_9 - h_1 h_6 h_8 \\ &\quad + h_2 h_6 h_7 - h_3 h_5 h_7 + h_3 h_4 h_8) \\ &= (h_4 - h_4)(S) = (h_4 - h_5) \cdot \det(H),\end{aligned} \tag{32}$$

and

$$\begin{aligned}k_4 &= (h_7 - h_8)(h_3 h_4 h_8 - h_3 h_5 h_7 - h_2 h_4 h_9 \\ &\quad + h_1 h_5 h_9 - h_1 h_6 h_8 + h_2 h_6 h_7) \\ &= (h_7 - h_8)(S) = (h_7 - h_8) \cdot \det(H),\end{aligned} \tag{33}$$

or

$$P_{v4} = (m_4, n_4, 1) = \left(\frac{h_1 - h_2}{h_7 - h_8}, \frac{h_4 - h_5}{h_7 - h_8}, 1\right). \quad (34)$$

*C. The equivalence of (2) and (3)*

To find the principal line according to the formulation elaborated in [9], equations of the four vanishing points, i.e., (15), (21), (28), and (34), can be substituted into (3) directly, or

$$\left(\frac{h_2 h_7 - h_1 h_8}{h_7 h_8}\right) u + \left(\frac{h_5 h_7 - h_4 h_8}{h_7 h_8}\right) v + c = 0, \quad (35)$$

with

$$c = -\left\{ \left(\frac{h_2}{h_8} - \frac{h_1}{h_7}\right) \frac{\left[\left(\frac{h_1}{h_7} \cdot \frac{h_2}{h_8}\right) - \left(\frac{h_1 + h_2}{h_7 + h_8}\right)\left(\frac{h_1 - h_2}{h_7 - h_8}\right)\right]}{\frac{h_1}{h_7} + \frac{h_2}{h_8} - \frac{h_1 + h_2}{h_7 + h_8} - \frac{h_1 - h_2}{h_7 - h_8}} \right.$$
$$\left. + \left(\frac{h_5}{h_8} - \frac{h_4}{h_7}\right) \frac{\left[\left(\frac{h_4}{h_7} \cdot \frac{h_5}{h_8}\right) - \left(\frac{h_4 + h_5}{h_7 + h_8}\right)\left(\frac{h_4 - h_5}{h_7 - h_8}\right)\right]}{\frac{h_1}{h_7} + \frac{h_2}{h_8} - \frac{h_1 + h_2}{h_7 + h_8} - \frac{h_1 - h_2}{h_7 - h_8}} \right\}.$$

$$= -\left\{\left(\frac{h_2 h_7 - h_1 h_8}{h_7 h_8}\right) \right.$$
$$\left[\frac{h_1 h_2 h_7{}^2 - h_1 h_2 h_8{}^2 - h_1{}^2 h_7 h_8 - h_2{}^2 h_7 h_8}{(h_2 h_7 - h_1 h_8)(h_7{}^2 + h_8{}^2)}\right]$$
$$+ \left(\frac{h_5 h_7 - h_4 h_8}{h_7 h_8}\right)$$
$$\left. \left[\frac{h_4 h_5 h_7{}^2 - h_4 h_5 h_8{}^2 - h_4{}^2 h_7 h_8 - h_5{}^2 h_7 h_8}{(h_5 h_7 - h_4 h_8)(h_7{}^2 + h_8{}^2)}\right]\right\}.$$

$$= -\frac{(h_2{}^2 + h_5{}^2 - h_1{}^2 - h_4{}^2) h_7 h_8 + (h_1 h_2 + h_4 h_5)(h_7{}^2 - h_8{}^2)}{h_7 h_8 (h_7{}^2 + h_8{}^2)}.$$
(36)

As $h_7 h_8 \neq 0$ is the condition for the existence of both (15) and (21), (35) is essentially the same as (2). Therefore, we can conclude that (2) and (3) are mathematically equivalent.

IV. EXTENSION FOR MORE GENERAL SITUATIONS

In this section, the equivalence of Equations (2) and (3) will be derived for more general situations which include: (i) the existence of an arbitrary orientation of the two sets of OVPs and (ii) the existence of *infinite* vanishing point. To that end, a simple way of deriving the OVPs is first presented.

*A. A simple and intuitive way of deriving OVPs*

Assume the plane containing the two squares shown in Fig. 2 is the X-Y plane in the 3D space. As $\overleftrightarrow{AB}$ and $\overleftrightarrow{DC}$ are parallel, their intersection can be expressed in homogeneous coordinate as $(1, 0, 0, 0)^T$. According to the projective geometry the projection of $(1, 0, 0, 0)^T$ on the image plane, i.e., $P_{v1}$, can be expressed as $\lambda_1 P_{v1} = K[R\ T](1, 0, 0, 0)^T$ based on a pinhole camera model with intrinsic matrix $K$, rotation matrix $R$ and translation vector $T$, and for some constant $\lambda_1$. Therefore, the corresponding vanishing point $P_{v1}$, and the OVP associated

with $\overleftrightarrow{AD}$ and $\overleftrightarrow{BC}$, or $P_{v2}$, can be expressed as

$$\lambda_1 P_{v1} = K[R\ T]\begin{bmatrix}1\\0\\0\\0\end{bmatrix} = K[r_1\ r_2\ r_3\ T]\begin{bmatrix}1\\0\\0\\0\end{bmatrix} = K[r_1\ r_2\ T]\begin{bmatrix}1\\0\\0\end{bmatrix}$$
$$= H\begin{bmatrix}1\\0\\0\end{bmatrix} = \begin{bmatrix}h_1 & h_2 & h_3\\h_4 & h_5 & h_6\\h_7 & h_8 & h_9\end{bmatrix}\begin{bmatrix}1\\0\\0\end{bmatrix} = \begin{bmatrix}h_1\\h_4\\h_7\end{bmatrix}, \quad (37)$$

and for some constant $\lambda_2$,

$$\lambda_2 P_{v2} = K[R\ T]\begin{bmatrix}0\\1\\0\\0\end{bmatrix} = K[r_1\ r_2\ r_3\ T]\begin{bmatrix}0\\1\\0\\0\end{bmatrix} = K[r_1\ r_2\ T]\begin{bmatrix}0\\1\\0\end{bmatrix}$$
$$= H\begin{bmatrix}0\\1\\0\end{bmatrix} = \begin{bmatrix}h_1 & h_2 & h_3\\h_4 & h_5 & h_6\\h_7 & h_8 & h_9\end{bmatrix}\begin{bmatrix}0\\1\\0\end{bmatrix} = \begin{bmatrix}h_2\\h_5\\h_8\end{bmatrix}, \quad (38)$$

respectively.

Similarly, the following expressions can be obtained for $P_{v3}$ and $P_{v4}$, respectively, as

$$\lambda_3 P_{v3} = H\begin{bmatrix}1\\1\\0\end{bmatrix} = \begin{bmatrix}h_1 & h_2 & h_3\\h_4 & h_5 & h_6\\h_7 & h_8 & h_9\end{bmatrix}\begin{bmatrix}1\\1\\0\end{bmatrix} = \begin{bmatrix}h_1 + h_2\\h_4 + h_5\\h_7 + h_8\end{bmatrix} \quad (39)$$

and

$$\lambda_4 P_{v4} = H\begin{bmatrix}-1\\1\\0\end{bmatrix} = \begin{bmatrix}h_1 & h_2 & h_3\\h_4 & h_5 & h_6\\h_7 & h_8 & h_9\end{bmatrix}\begin{bmatrix}-1\\1\\0\end{bmatrix} = \begin{bmatrix}-h_1 + h_2\\-h_4 + h_5\\-h_7 + h_8\end{bmatrix}. \quad (40)$$

It is easy to see that for finite OVPs, (37) to (40) can be directly rewritten in form of their counterparts in Sec. III, i.e., (15), (21), (28), and (34).

*B. PL for arbitrarily oriented two sets of OVPs*

In Sec. III, OVPs $P_{v1}$, $P_{v2}$, $P_{v3}$ and $P_{v4}$ are considered for equally spaced directions of 0°, 90°, 45°, and 135° obtained the two unit squares shown in Fig. 2. Without loss of generality, assume one set of OVPs correspond directions of 0° and 90°, we can use the rectangle shown in Fig. 3, in place of Square I in Fig. 2, to extend our derivation to the more general case wherein directions of the second set of OVPs do not correspond to 45° or 135°. While there will be no change to $P_{v1}$ and $P_{v2}$, new expressions of for $P_{v3}$ and $P_{v4}$ can be obtained, similar to that done in (39) and (40), as

$$\lambda_3 P_{v3} = H \begin{bmatrix} a \\ b \\ 0 \end{bmatrix} = \begin{bmatrix} h_1 & h_2 & h_3 \\ h_4 & h_5 & h_6 \\ h_7 & h_8 & h_9 \end{bmatrix} \begin{bmatrix} a \\ b \\ 0 \end{bmatrix}$$
$$= \begin{bmatrix} ah_1 + bh_2 \\ ah_4 + bh_5 \\ ah_7 + bh_8 \end{bmatrix} \quad (41)$$

$$\lambda_4 P_{v4} = H \begin{bmatrix} -b \\ a \\ 0 \end{bmatrix} = \begin{bmatrix} h_1 & h_2 & h_3 \\ h_4 & h_5 & h_6 \\ h_7 & h_8 & h_9 \end{bmatrix} \begin{bmatrix} -b \\ a \\ 0 \end{bmatrix}$$
$$= \begin{bmatrix} -bh_1 + ah_2 \\ -bh_4 + ah_5 \\ -bh_7 + ah_8 \end{bmatrix}. \quad (42)$$

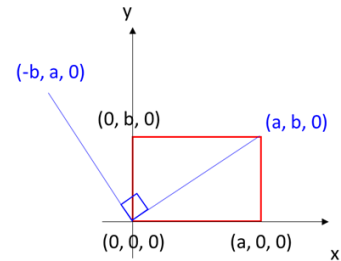

**Figure 3.** A rectangle pattern with a new set of OVPs.

By substituting $P_{v1}$ to $P_{v4}$ into (3), with new expressions of

$$P_{v3} = \left(\frac{ah_1 + bh_2}{ah_7 + bh_8}, \frac{ah_4 + bh_5}{ah_7 + bh_8}, 1\right)$$
$$P_{v4} = \left(\frac{-bh_1 + ah_2}{-bh_7 + ah_8}, \frac{-bh_4 + ah_5}{-bh_7 + ah_8}, 1\right), \quad (43)$$

the line equation of PL will have no change in the first two coefficients, i.e.,

$$\left(\frac{h_2 h_7 - h_1 h_8}{h_7 h_8}\right) u + \left(\frac{h_5 h_7 - h_4 h_8}{h_7 h_8}\right) v + c = 0, \quad (44)$$

with

$$c = -\left\{\left(\frac{h_2}{h_8} - \frac{h_1}{h_7}\right) \frac{\left[\left(\frac{h_1}{h_7} \cdot \frac{h_2}{h_8}\right) - \left(\frac{ah_1 + bh_2}{ah_7 + bh_8}\right)\left(\frac{-bh_1 + ah_2}{-bh_7 + ah_8}\right)\right]}{\frac{h_1}{h_7} + \frac{h_2}{h_8} - \left(\frac{ah_1 + bh_2}{ah_7 + bh_8}\right) - \left(\frac{-bh_1 + ah_2}{-bh_7 + ah_8}\right)}\right.$$
$$+ \left(\frac{h_5}{h_8} - \frac{h_4}{h_7}\right) \frac{\left[\left(\frac{h_4}{h_7} \cdot \frac{h_5}{h_8}\right) - \left(\frac{ah_4 + bh_5}{ah_7 + bh_8}\right)\left(\frac{-bh_4 + ah_5}{-bh_7 + ah_8}\right)\right]}{\frac{h_4}{h_7} + \frac{h_5}{h_8} - \left(\frac{ah_4 + bh_5}{ah_7 + bh_8}\right) - \left(\frac{-bh_4 + ah_5}{-bh_7 + ah_8}\right)}\right\}.$$

$$= -\left\{\left(\frac{h_2 h_7 - h_1 h_8}{h_7 h_8}\right)\right.$$
$$\left[\frac{-ab(h_1 h_2 h_7{}^2 - h_1 h_2 h_8{}^2 - h_1{}^2 h_7 h_8 - h_2{}^2 h_7 h_8)}{-ab(h_2 h_7 - h_1 h_8)(h_7{}^2 + h_8{}^2)}\right]$$
$$+ \left(\frac{h_5 h_7 - h_4 h_8}{h_7 h_8}\right)$$
$$\left.\left[\frac{-ab(h_4 h_5 h_7{}^2 - h_4 h_5 h_8{}^2 - h_4{}^2 h_7 h_8 - h_5{}^2 h_7 h_8)}{-ab(h_5 h_7 - h_4 h_8)(h_7{}^2 + h_8{}^2)}\right]\right\}.$$

$$= -\frac{(h_2{}^2 + h_5{}^2 - h_1{}^2 - h_4{}^2) h_7 h_8 + (h_1 h_2 + h_4 h_5)(h_7{}^2 - h_8{}^2)}{h_7 h_8 (h_7{}^2 + h_8{}^2)},$$

which is exactly the same as the constant in (2).

*C. PL for infinite vanishing point*

While no constraint on location of any member of the two sets of OVPs is given in [7], location at infinity is implicitly excluded in the derivation of PL presented in [9]. In this subsection, a fairly simple extension of the derivation of PL for $P_{v1}$ located at infinity is provided, while similar extensions for other OVPs are omitted for brevity.

For $P_{v1}$ located at infinity, e.g., with $\overleftrightarrow{A'B'} \parallel \overleftrightarrow{C'D'}$ for the image of Square I in Fig. 2, its homogeneous coordinate will become

$$P_{v1} = [m_1 \ n_1 \ 0]^T. \quad (45)$$

By taking the limit of the PL expression (3) for arbitrarily large $m_1$ and $n_1$, we will have

$$-m_1 u + (-n_1) v + c = 0, \quad (46)$$

with

$$c = -[-m_1 m_2 - n_1 n_2] \quad (47)$$

As the PL expression in (46) is much simpler than that for (3), a formal proof of its equivalence with (2) is omitted for brevity.

V. CONCLUSION

In this report, an algebraic proof of the equivalence of two PL expressions, either derived in [7] in terms of elements of the homography matrix or established in [9] for two sets of orthogonal vanishing points, is provided. Accordingly, four orthogonal vanishing points are firstly derived for a given homography matrix, wherein two unit square are employed to simplify the derivation. Then, the PL for these vanishing points, which has exactly the same line equation as that derived in [7], is obtained with the procedure outlined in [9]. Moreover, with another simple and intuitive way of deriving orthogonal vanishing points, the foregoing equivalence of PL expression is considered for more general situations which include: (i) the existence of arbitrary orientations of the two sets of OVPs, i.e., other than orientations associated with squares and their diagonals, and (ii) the existence of *infinite* vanishing point.